\documentclass[10pt,twocolumn,letterpaper]{article}

\usepackage{cvpr}
\usepackage{times}
\usepackage{epsfig}
\usepackage{graphicx}
\usepackage{amsmath}
\usepackage{amssymb}
\usepackage{multirow}
\usepackage{floatrow}
\usepackage{tabularx}
\usepackage{array}
\newfloatcommand{capbtabbox}{table}[][\Fbwidth]
\usepackage{booktabs}
\usepackage{arydshln} 
\usepackage{subcaption}

\usepackage[breaklinks=true,bookmarks=false]{hyperref}

\cvprfinalcopy 

\def\cvprPaperID{****} 

\begin{document}

\title{Sub-clusters of Normal Data for Anomaly Detection}

\author{Gahye Lee, Seungkyu Lee \\
The college of Electronics and Information, Kyung hee University, Republic of Korea\\
{\tt\small \{waldstein94, seungkyu\}@khu.ac.kr}}

\maketitle

\begin{abstract}
   Anomaly detection in data analysis is an interesting but still challenging research topic in real world applications. As the complexity of data dimension increases, it requires to understand the semantic contexts in its description for effective anomaly characterization. However, existing anomaly detection methods show limited performances with high dimensional data such as ImageNet.
    Existing studies have evaluated their performance on low dimensional, clean and well separated data set such as MNIST and CIFAR-10.
    In this paper, we study anomaly detection with high dimensional and complex normal data. Our observation is that, in general, anomaly data is defined by semantically explainable features which are able to be used in defining semantic sub-clusters of normal data as well. 
    We hypothesize that if there exists reasonably good feature space semantically separating sub-clusters of given normal data, unseen  anomaly also can be well distinguished in the space from the normal data. 
    We propose to perform semantic clustering on given normal data and train a classifier to learn the discriminative feature space where anomaly detection is finally performed.
    Based on our careful and extensive experimental evaluations with MNIST, CIFAR-10, and ImageNet with various combinations of normal and anomaly data, we show that our anomaly detection scheme outperforms state of the art methods especially with high dimensional real world images.  
\end{abstract}

\section{Introduction}
Anomaly detection is identification task of unusual samples that have hardly been observed from majority of normal data. "Unusual sample" can be semantically defined at each application such as abnormal behavior detection, disease detection, outlier in signal, unexpected situation, etc \cite{motion}, \cite{network}, \cite{gan2017backprop}.
There has been a large body of research work on anomaly detection such as learning representation of normal distribution, using deep generative network and pre-trained classifier, which are well-categorized and summarized in Chalapathy et al. \cite{survey2019deep} and Chandola et al. \cite{survey2009anomaly}.

\begin{figure}[!t] 
  \centering
  \includegraphics[width=1\linewidth]{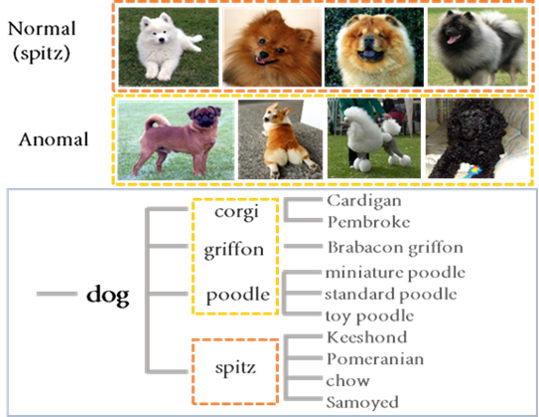} 
  \caption{Examples of semantic clusters in normal data and anomalies. When the 'spitz' is normal, its semantic clusters could be sub-classes such as 'Keeshond', 'Pomeranian', 'chow', 'Samoyed' and sub-classes in 'corgi', 'griffon', 'poodle' could be anomalous samples. In this paper, our final goal is distinguishing semantic anomalous samples.} 
  \label{hierarchical}
\end{figure} 
\textbf{Representation in Latent Space}
A lot of anomaly detection methods based on statistical approaches have been proposed focusing on the optimal description of normal data in a latent space maximizing anomaly detection \cite{markou2003novelty}, \cite{anomaly2009survey}. In Eskin and Eleazar \cite{noise2000probability}, Zong et al.  \cite{gmm}, Yamanishi et al. \cite{yamanishi2004line}, methods estimate the probability distribution of normal data using mixture models and evaluate the likelihood of each sample to be considered as an anomaly. There is a non-parametric approach in estimating density of data \cite{latent2019autoregressive}. They build a deep auto-encoder with parametric density estimator and learn statistical distribution of normal data in the latent space via an auto-regressive procedure. 
On the other hand, there is an approach that concentrates on finding a feature space having regular factors of normal distribution. Ruff et al. \cite{deeponeclasss} inspired by kernel-based one-class classification method learn a transformation into hyper-sphere minimizing its volume. This task forces trained network to extract the common factors of normal data so that any anomaly sample falls outside of normal distribution territory. 
 
\textbf{Employing Generative Adversarial Networks}
Since GANs\cite{goodfellow2014generative} was introduced that learns mapping from a latent space to image space, there have been various adversarial models solving anomaly detection problem \cite{perera2019ocgan}, \cite{adversarial}, \cite{zenati2018efficient}, \cite{gan2017backprop}.
Perera et al. \cite{perera2019ocgan} train an auto-encoder for constrained latent representations of given class with discriminator of GANs and make the decoder map the bounded space to the image space. 
Generator of GANs is used for enhancing in-liers and distorting out-liers for better decision with original samples\cite{adversarial}. Schlegl et al. \cite{gan2017backprop}, inversely map the query sample to a latent space using pre-trained GANs and find a closest point which is visually the most similar instance to query sample for reconstruction.

\textbf{Auto-encoder}
Using Auto-encoder, carrying out the reconstruction of input is also popular approach for learning representations. For example, Cong et al. \cite{recon1} propose new criteria, sparse reconstruction cost, and measure the normality over the test samples. Zhou and Paffenroth \cite{anomaly2017autoencoder} propose a deep auto-encoder that increases the ability of anomaly detection based on robust PCA. However, using auto-encoder network sometimes 'generalizes' itself on both normal and anomalous data leading to miss detection of anomalies \cite{gong2019memorizing}. To alleviate the problem, the authors suggest an augmented auto-encoder with a memory module (MemAE) which reconstructs query input via retrieving the most relevant memory items made from normal dataset. 

\textbf{Out-of-Distribution Detection}
Trained classifier on real-image through classification task tends to provide high-confidence value and fails to detect out-of-distribution examples in Hendrycks et al. \cite{msp}. They propose a detection baseline based on a discovered phenomenon that the prediction probability of in-distribution is higher than the prediction probability of out-of-distribution. Liang et al. \cite{odin} propose that in and out-of-distribution can be more separated when temperature scaling is used and small perturbations through gradient ascent is added to the input. 

\textbf{Self-Supervised Learning}
In computer vision, supervised learning have successfully presented outstanding performance in many tasks such as object detection and classification, scene segmentation and so on \cite{badrinarayanan2017segnet}, \cite{yolo}, \cite{resnet}. 
Self-supervised learning is kind of representation learning which requires only pretext learning task such as predicting the direction of rotation, relationships in an image \cite{kolesnikov2019revisiting}, \cite{rot2018unsupervised}, \cite{jigsaw}. 
It learns semantic characteristics of unlabeled data, that helps anomaly detection tasks \cite{geo2018transform}, \cite{neighborssl}, \cite{self2019supervised}, \cite{selfsupervised2018ensemble}, \cite{anomaly2017autoencoder}.  
In general, the scope and characteristics of such novel instances are not able to be explicitly defined and training a classification network for detecting anomaly samples are not a trivial task. Thus, characterizing and detecting all the possible unseen instances might be an ill-posed problem. However, in many researches, attempts to detect anomalies whose range is considered very wide and ambiguous are very common.
    
\begin{figure}[!t] 
      \centering
      \includegraphics[width=1\linewidth]{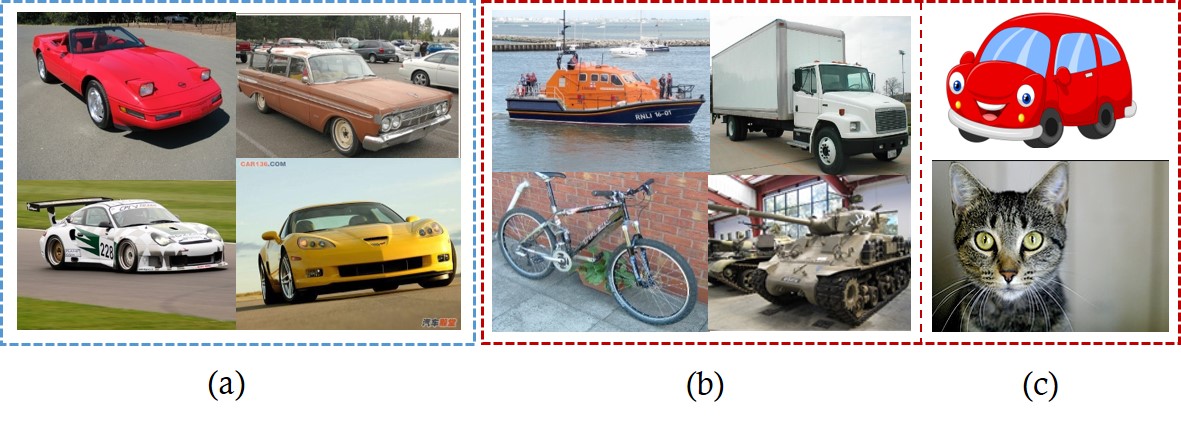} 
      \caption{Examples of semantic anomalies (a) Normal samples of 'car' class (b) 'semantic anomalies' (c) semantically less relevant anomaly samples} 
      \label{intro}
    \end{figure}

When normal data is high-dimensional and complex, such as an high-resolution image with multiple objects, explicitly defining the scope of 'anomalous' becomes complicated task. Our observation in anomaly detection is that, in general, normality of normal data implicitly includes the characteristic of contextual aspects of given majorities. Therefore, the distinction between normal and anomaly varies along the semantically common aspects of normal samples.
Rather than pursuing statistically optimal method over all anomaly detection tasks that is an insoluble puzzle, we endeavor to find semantically explainable features of unseen anomaly cases within normal samples.
To this end, we narrow down the scope of anomaly based on semantic information of normal data. For example, with image data, explainable characteristics such as object types, location, weather, appeared person in an image are considered as criteria defining anomaly. 
In other words, in general, anomaly is able to be predictable from semantic understanding of normal data (that is semantic anomaly). Note that we need to learn the separation of anomaly from normal, not the distribution itself or entire characteristics of anomaly data. Here is examples in figure \ref{intro}. When the set of car images is given as normal, semantic anomalies related to the normal data could be other transportations such as bicycles, tanks, trucks or ships. However, semantically less relevant samples from farther object categories such as cartoon cars and cats are hardly be actual anomalies.
    
Similar studies have been introduced in prior work. In the surveys such as Chandola et al. \cite{survey2009anomaly}, Chalapath et al. \cite{survey2019deep}, anomalies are categorized by two groups, simple and complex anomalies. The latter is defined as contextual and collective anomalies. In this case, they termed a data instance that is anomalous in a specific context as 'contextual anomalies', which is quite close to what we are interested in. In traffic analytic application \cite{traffic}, they suggest detection algorithms based on expected anomalous characteristics of vehicles such as immobile vehicles for longer time than other normal vehicles or significantly slower vehicles than surrounding traffic. 

In this paper, we propose 1) an anomaly detection framework learning sub-clusters of normal data that provides semantic feature space for explainable and improved anomaly detection, 2) an extensive anomaly detection evaluation tests with MNIST, CIFAR-10, and ImageNet, providing comprehensive evaluation results and analysis, where our semantic anomaly detection shows outperforming performance in both quantitative and qualitative manners.  

\section{Problem Statement} 
\begin{figure}[!t] 
  \centering
  \includegraphics[width=1\linewidth]{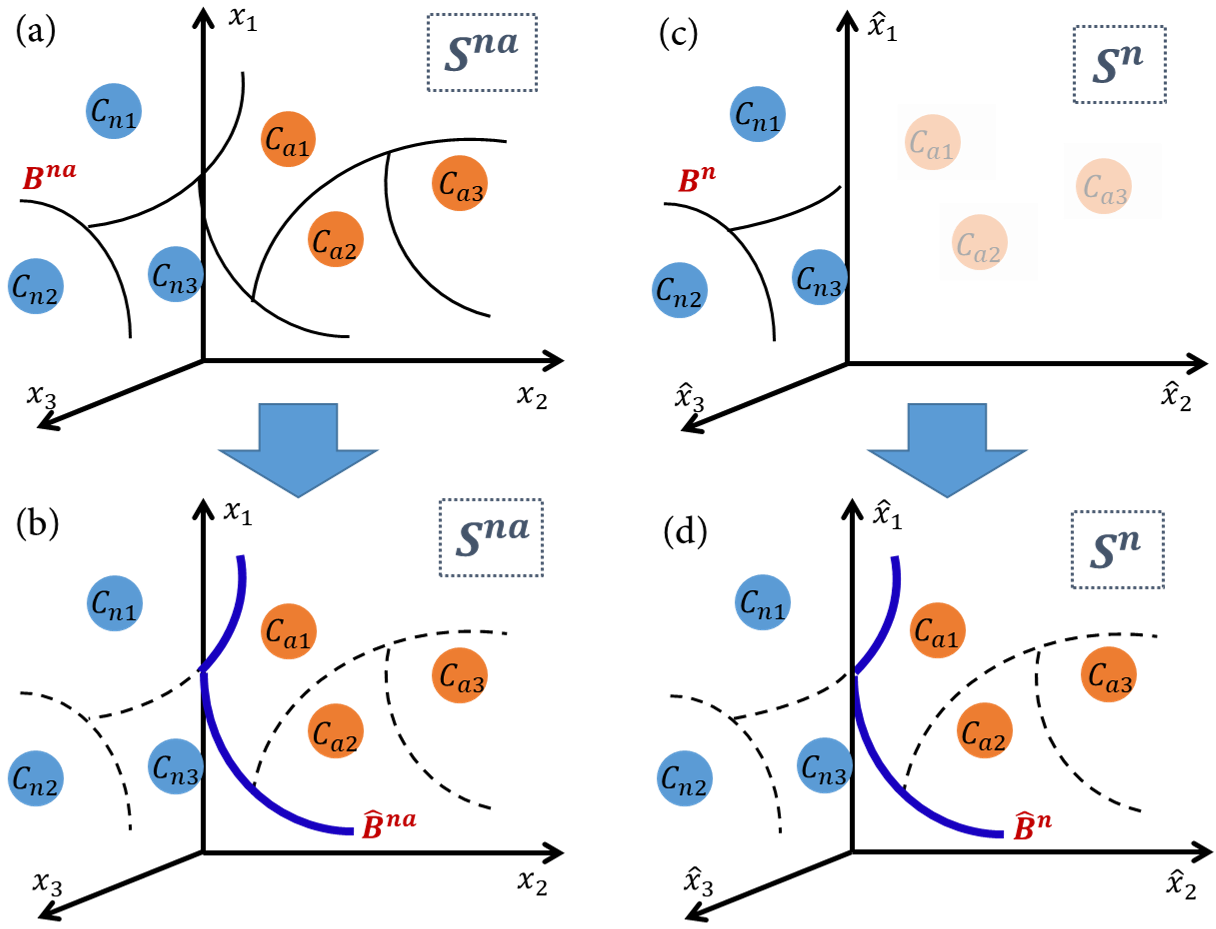} 
  \caption{The concept of our hypothesis, separability of feature set on ideal and real situation. First two sub-figures show a ideal simulation when using both normal and anomal set. Next two sub-figures show our assumed situation when using only normal set. Anomalies would be separated in certain space that learns semantic properties of sub-clusters in normal data.}  
  \label{concept}
\end{figure}

Let normal and anomaly data have respective several sub-classes and all sub-classes from both normal and anomaly are well separated by certain semantically explainable feature set $S^{na}$ where $B^{na}$ is a set of decision boundaries for all sub-classes (See the Figure \ref{concept} (a). Sub-clusters of normal and anomal data are represented as $C_{ni}$, $C_{ai}$ respectively where $i$ is $1, 2, 3$.). Then the feature set $S^{na}$ is also able to separate normal and anomaly data very well with a decision boundary $\widehat{B^{na}}$ that is the subset of $B^{na}$ (Figure \ref{concept} (b)). 
Now assume that we only use sub-classes of normal data finding new feature set $S^n$ providing good separation for normal sub-classes with the set of decision boundaries $B^n$ (Figure \ref{concept} (c)). 
As we previously have discussed, in real applications, if all sub-classes are separated by semantically meaningful aspects, we assume that two feature sets $S^{na}$ and $S^n$ have to be similar to each other with similar semantic meaning ($S^{na} \approx S^n$).
As a result, we assume that feature set $S^n$ is also able to separate normal and anomaly data very well where ($\widehat{B^{na}} \approx \widehat{B^n}$) (Figure \ref{concept} (d)).
Therefore, if there is a good feature space separating semantic sub-clusters of given normal data, unseen semantic anomaly also can be distinguished well from entire normal data in that space. 
For example, in hierarchical tree (top) which is part of ImageNet shown in figure \ref{hierarchical}, let us define spitz as a normal data and corgi, griffon, poodle as an anomaly data. In this case, four sub-classes (Samoyed, Pomeranian, chow, keeshond) of spitz are kind of semantically explainable sub-clusters.  
If we find a good classifier distinguishing sub-clusters of spitz, detection scores of the classifier of test samples provide normal and anomaly separation.

\begin{figure}[!b]
    \centering  
        \includegraphics[width=0.32\linewidth]{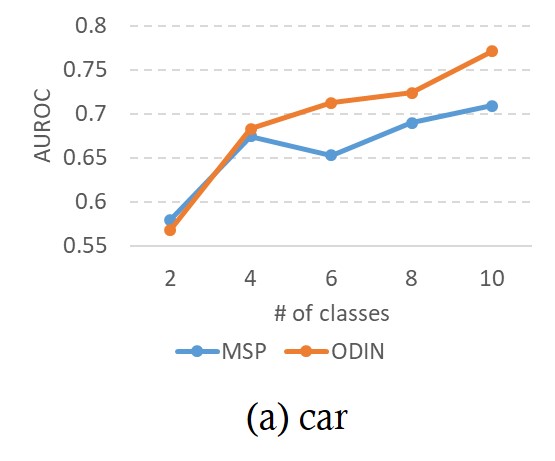} 
        \includegraphics[width=0.32\linewidth]{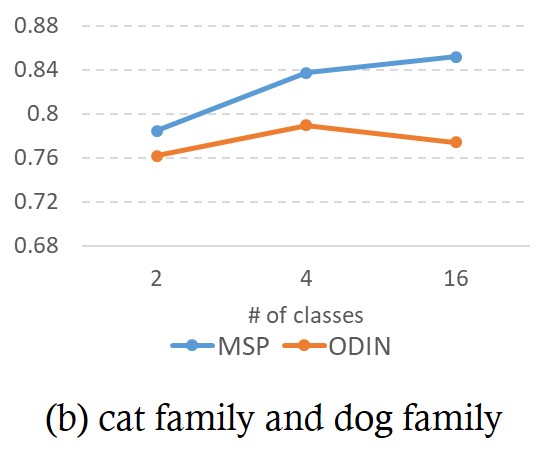}  
        \includegraphics[width=0.32\linewidth]{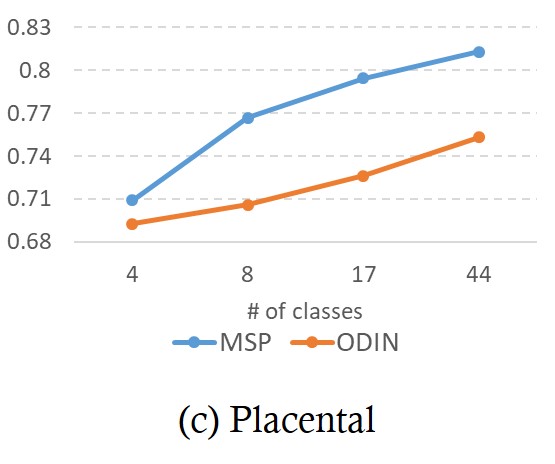}\\
         
    \caption{AUROC score corresponding to the number of clusters in each dataset. To find the better feature set for anomaly detection, for (a) ’car’, (b) 'cat family and dog family', (c) 'Placental' synset in table 10, we incrementally increase the number of cluster in normal data such as 2-4-6-8-10, 2-4-16 and 4-8-17-44. Y-axis and X-axis depict the AUROC score and the number of clusters respectively. For calculating AUROC value, we use MSP and ODIN which are represented by blue and orange line}
    \label{nclutsers}
\end{figure}

    In order to verify our assumption, we perform simple tests with 'car', 'cat family and dog family' and 'placental' image sets of ImageNet. 'car' has 10 sub-classes and other types of transportation are used as its anomaly. 'cat family and dog family' has 16 sub-classes and other categories of animal are used as its anomaly. 'placental' set has 44 sub-classes and sub-classes in 'metatheria' are used as its anomaly. For example, a 'car' set having ten sub-classes is assigned to normal and other related classes such as golf-cart, horse cart, mountain bike are assigned to anomaly data. In our simple anomaly detection tests, we increase the number of clusters in normal from two to ten. Note that, in this test, each cluster consists of labeled sub-classes. To be specific, we divide normal car set into two clusters having five sub-classes at each cluster. Next, normal is divided into four clusters, and so on. We assume that the labeled classes are good examples of semantic clusters.
    Graphs in figure \ref{nclutsers} and table \ref{incr_comb} (a),(b),(c) represent the above simple test results by AUROC score showing that if normal data is semantically clustered more, anomaly detection score increases. 
This simple test presents the ability of sub-clustering of normal data in anomaly detection.
If we have more number of clusters for identical normal data, classification process finds better feature sets separating the clusters to each other. As a result, anomalous samples with respect to semantically similar descriptions are also well separated from all sub-clusters of normal data. 
    
\begin{figure*}[t] 
  \centering
  \includegraphics[width=1\linewidth]{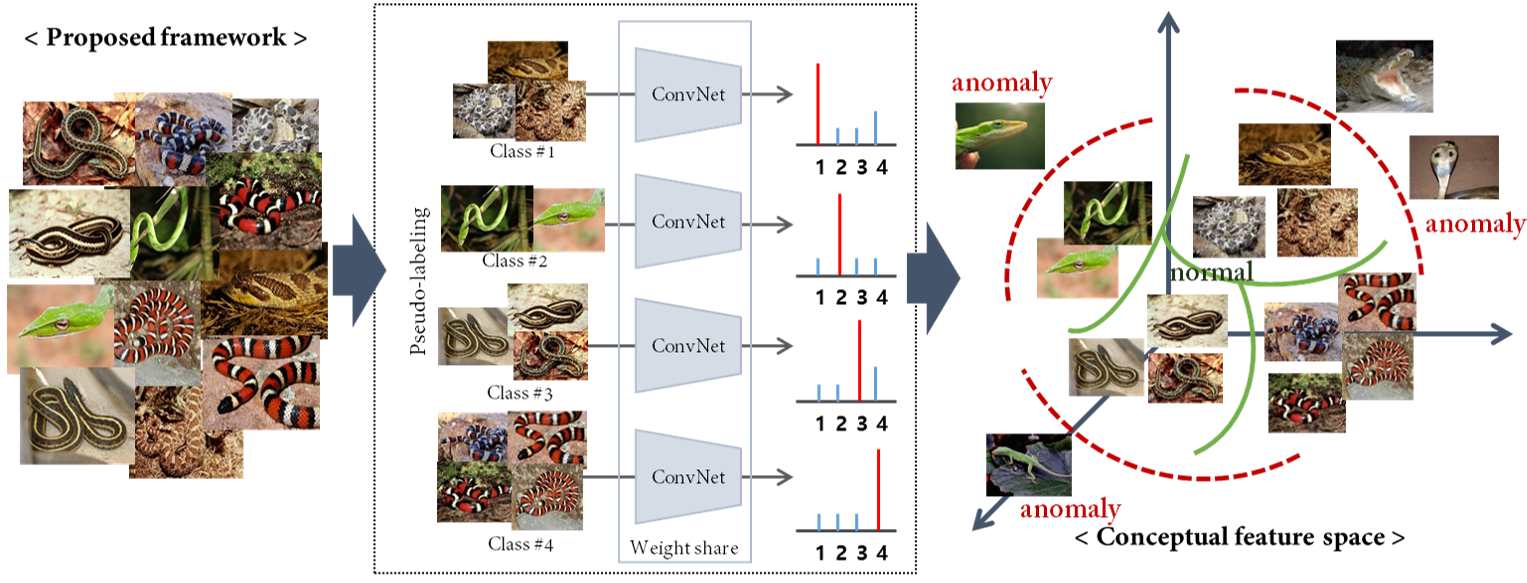} 
  \caption{Diagram of proposed method. In this case, we use 'snake' data as normal and divide normal into 4 clusters based on class label. Each cluster given pseudo-label is used for training classifier. Expected feature space gives good separation not just among normal sub-clusters but also semantic anomaly} 
  \label{processing}
\end{figure*}

    Moreover, we assume that the type of semantic property separating normal data clusters is critical in anomaly detection.
    In this simple test, labeled dataset like Imagenet already has its semantic criterion 'object'. Instead, we make four normal data clusters at four different object levels respectively. These four different semantic clusterings come to have different semantic property (different object classification level).
    Using 'placental' set, we build four clusters and test the separability of anomaly data for each case. Result is shown in table \ref{incr_comb} (d) at four different levels.
    As the result shows, highest AUROC score is obtained at level-2 and scores decrease as the level goes down away from the original normal data class level. 
    This test shows that there exists an optimal level in normal data clustering in which probably we can get the most semantically meaningful separation in anomaly detection.
Through the simple tests, we study how anomalous data is well separated from normal data depending on the number of clusters and the type of clusters.
We assume that when there exist a set of semantic clusters of a normal set and if we can find feature space that distinguishes sub-clusters, semantic anomaly data also can be identified in the same feature space. 
Figure \ref{processing} summarizes our proposed method for anomaly detection. 

\begin{table*} 
            \centering
            \captionsetup[subtable]{position = below}
            \caption{The AUROC scores of 'car', 'cat family and dog family' and 'placental' synset. AUROC score increases  as we increase the number of sub-cluster in normal data. In (d) we test the separation ability according to different semantic property (different object classification level).} \label{incr_comb}
            \begin{subtable}{0.22\linewidth}
            \centering
            {\scriptsize
            \begin{tabular}{ccc} 
            \toprule
            N(cluster)&  MAX &	ODIN    \\ 
            \hline
            2& 	0.58 & 0.57   \\
            4 & 0.67 &	0.68 \\
            6 & 0.65 &	0.71  \\
            8 & 0.69&	0.72 \\
            10 & 0.71 &	0.77 \\
            \toprule
            \end{tabular}}
            \caption{'car'}
          \end{subtable}%
          \hspace*{1em}
          \begin{subtable}{0.22\linewidth}
            \centering
            {\scriptsize
            \begin{tabular}{ccc} 
            \toprule
            N(cluster)&  MAX &	ODIN    \\ 
            \hline
            2& 	0.78 & 0.76   \\
            4& 0.84 &	0.79 \\
            16& 0.85 &0.77  \\ 
            \toprule
            \end{tabular}
            }
            \caption{'cat family and dog family'}
          \end{subtable}%
          \hspace*{1em}  
          \begin{subtable}{0.22\linewidth}
            \centering
            {\scriptsize
            \begin{tabular}{ccc} 
            \toprule
            N(class)&  MAX &	ODIN    \\ 
            \hline
            4& 	0.71 & 0.69   \\
            8& 0.77 &	0.71 \\
            17& 0.79 &0.73  \\ 
            44& 0.81 &0.75  \\ 
            \toprule
            \end{tabular}
            \caption{'placental'}}
          \end{subtable}%
          \hspace*{1em}  
          \begin{subtable}{0.22\linewidth}
          \centering
          {\scriptsize
          \begin{tabular}{ccc} 
            \toprule
              Class-level  &  MAX &	ODIN    \\ 
            \hline
            level-1 & 	0.71 & 0.62   \\
            level-2 & \textbf{0.79} &	0.73 \\
            level-3 & 0.76 &	0.62  \\
            level-4  & 0.75&	0.62 \\ 
                \toprule
                \end{tabular}}
                \caption{'placental'}
          \end{subtable}
\end{table*}

 \subsection{Methodology}  
Given normal data $X= \{ x_1, x_2, ..., x_n \}$ of n samples, we divide the samples into k multiple clusters, $X_i = \{x_1,x_2,.. ., x_m\}$, then $X = X_1 \cup X_2 \cup...\cup X_k$. We assign pseudo-labels $y_i$ to each cluster $X_i$. After mapping the label, we train a classifier $f_\theta(x)$ to find a $\theta^*$ which makes good separations among the clusters $X_i$.

Once classifier finishes learning, detection score is calculated by : 
\begin{equation}
  S(x, f_\theta)=\begin{cases}
    Normal \; Class, & \text{if $S(x,f_\theta)>\gamma $}.\\
    Anomaly, & \text{otherwise}.
  \end{cases}
\end{equation}
with pre-defined threshold $\gamma$. For the score function $S(x)$, we follow two algorithms\cite{msp} and \cite{odin} noted as MAX and ODIN respectively(explained in detail in section \ref{expresults}) 
For the evaluation of proposed method, semantic clustering has to be performed on unlabeled normal data, however, developing a clustering method optimal for a particular anomaly detection is out of scope of this work.
Instead, in this paper, we demonstrate performance of proposed method through label information of dataset and visual context learning method. In the first type of experiment, sub-classes of normal dataset are considered as contextual sub-clusters and we learn the semantic feature space of sub-classes. In the second type of experiment, we learn the semantic feature space of unlabeled data using existing method which learns its visual similarity.

\begin{table*}[!t]
\begin{center}
    \caption{Test 1 : AUROC scores of previous works on MNIST data. Normal set is composed of single class, anomaly is composed of multiple classes. }
\label{prevmnist}
\begin{tabular}{crrrrrrrrrrr} %
\toprule
        &0 &1&2&3 &4&5 &6&7 &8 &9  & mean \\
\hline 
OCSVM&	\textbf{0.988}&	\textbf{0.999}&	0.902&	0.95&	0.955&	0.968&	0.978&	0.965&	0.853&	0.955&	0.951\\
KDE&	0.885&	0.996&	0.71&	0.693&	0.844&	0.776&	0.861&	0.884&	0.669&	0.825&	0.814\\
DAE&	0.894&	\textbf{0.999}&	0.792&	0.851&	0.888&	0.819&	0.944&	0.922&	0.74&	0.917&	0.877\\
VAE&	0.997&	\textbf{0.999}&	0.936&	0.959&	0.973&	0.964&\textbf{0.993}&	\textbf{0.976}&	0.923&	0.976&	0.97\\
Pix CNN&0.531&	0.995&	0.476&	0.517&	0.739&	0.542&	0.592&	0.789&	0.34&	0.662&	0.618\\
GAN\cite{goodfellow2014generative}&	0.926&	0.995&	0.805&	0.818&	0.823&	0.803&	0.89&	0.898&	0.817&	0.887&	0.866\\
AND&	0.984&	0.995&	0.947&	0.952&	0.96&	0.971&	0.991&	0.97&	0.922&	0.979&	0.967\\
AnoGAN\cite{gan2017backprop}&	0.966&	0.992&	0.85&	0.887&	0.894&	0.883&	0.947&	0.935&	0.849&	0.924&	0.913\\
DSVDD&	0.98&	0.997&	0.917&	0.919&	0.949&	0.885&	0.983&	0.946&	\textbf{0.939}&	0.965&	0.948\\
OCGAN\cite{perera2019ocgan}&	0.998&	\textbf{0.999}&	\textbf{0.942}&\textbf{	0.963}&	\textbf{0.975}&	\textbf{0.98}&	0.991&	0.981&	\textbf{0.939}&	\textbf{0.981}&	\textbf{0.975}\\

Jigsaw &	0.983&	0.838&	0.847&	0.788&	0.855&	0.89&	0.973&	0.909&	0.884&	0.92&	0.889\\ 
\toprule
\end{tabular}
\end{center}
\end{table*}

\begin{table*}[!t]
\begin{center}
    \caption{Test 1 : AUROC scores of previous works on CIFAR-10 data. Normal set is composed of single class, anomaly is composed of multiple classes.}
\label{prevcifar}
\begin{tabular}{crrrrrrrrrrr}  
\toprule
        &PLANE &CAR&BIRD&CAT &DEER&DOG &GROG&HORSE &SHIP &TRUCK  & mean \\
\hline 
OCSVM   &0.63	&0.44   &	0.649   &	0.487   &	0.735   &	0.5     &	0.725   &	0.533   &	0.649   &	0.508&	0.586 \\
KDE     &0.658	&0.52   &	0.657   &	0.497   &	0.727   &	0.496   &	0.758   &	0.564   &	0.68    &	0.54&	0.61\\ 
DAE     &0.411	&0.478  &	0.616   &	0.562   &	0.728   &	0.513   &	0.688   &	0.497   &	0.487   &	0.378&	0.536\\
VAE	    &0.7	&0.386  &	0.679   &	0.535   &	0.748   &	0.523   &	0.687   &	0.493   &	0.696   &	0.386&	0.583\\
PixCNN	&0.788  &0.428  &   0.617   &	0.574   &	0.511   &	0.571   &	0.422   &	0.454   &	0.715   &	0.426&	0.551\\
GAN\cite{goodfellow2014generative}	    &0.708  &0.458  &	0.664   &	0.51    &   0.722   &	0.505   &	0.707   &	0.471   &	0.713   &	0.458&	0.592\\
AND	    &0.717  &0.494  &	0.662   &	0.527   &	0.736   &	0.504   &	0.726   &	0.56    &	0.68    &	0.566&	0.617\\
AnoGAN\cite{gan2017backprop}  &0.671  &0.547  &	0.529   &	0.545   &	0.651   &	0.603   &	0.585   &	0.625   &	0.758   &	0.665&	0.618\\
DSVDD   &0.617  &0.659  &	0.508   &	0.591   &	0.609   &	0.657   &	0.677   &	0.673   &	0.759   &	0.731&	0.648\\
OCGAN\cite{perera2019ocgan}   &0.757  &0.531  &	0.64    &	0.62    &	0.723   &	0.62    &	0.723   &	0.575   &	0.82    &	0.554&	0.657\\
Geometric\cite{geo2018transform} &0.762 & 0.848 & 0.771 & 0.732 & \textbf{0.828 }& 0.848 & 0.82& \textbf{0.887} & 0.895& 0.834 & 0.823\\
SSL\cite{self2019supervised} &\textbf{0.877}& \textbf{0.939} & \textbf{0.786} &\textbf{ 0.799} & 0.817 &\textbf{ 0.856 }& \textbf{0.933} & 0.879 & \textbf{0.926 }& \textbf{0.921} & \textbf{0.873}\\
Jigsaw&0.747	&0.828	&   0.638	&   0.562	&   0.637	&   0.685	&   0.682	&   0.774	&   0.764	&   0.774&	0.709\\ 
\toprule
\end{tabular}
\end{center}
\end{table*} 
 
\begin{table}[b]
\begin{center}
    \caption{Proposed anomaly detection test using MNIST, CIFAR-10, CIFAR-100 ImagetNet dataset}
\label{expsetting} 
{\scriptsize
\begin{tabular}{cccc} %
\toprule
\multirow{2}{*}{\shortstack{Traditional\\Evaluation}} & \multirow{2}{*}{Test 1} &  Normal of Single Class  &  MNIST\\ 
                        & & Anomaly of Multiple Classes   & CIFAR-10  \\
\hline 
\multirow{8}{*}{\shortstack{Proposed\\Evaluation}} &  \multirow{2}{*}{Test 2}  & Normal of Multiple Classes  &  MNIST \\
                    &    &  Anomaly of Multiple Classes & CIFAR-10  \\ \cline{2-4}
                    &  \multirow{3}{*}{Test 3}  & \multirow{3}{*}{ \shortstack{Normal of Multiple classes \\Anomal of Single Class}} & MNIST  \\
                     &   &      &	CIFAR-10 \\
                    &    &                   &	ImageNet \\ \cline{2-4}
                     & Test 4 & Semantic Clustering with Labeled data & ImageNet \\ 
                     & \multirow{2}{*}{Test 5} & \multirow{2}{*}{Semantic Clustering with Unlabeled data}  & \multirow{2}{*}{\shortstack{ImageNet\\ CIFAR-100}} \\ 

\\\toprule
\end{tabular} 
}
\end{center}
\end{table}

\begin{table*}[!t]
           \centering
           \captionsetup[subtable]{position = below}
          \captionsetup[table]{position=top}
           \caption{ Test2: AUROC scores corresponding to the number of sub-classes included by normal set. (a), (b) For MNIST and CIFAR-10 dataset, AUROC score is calculated as increasing the number of sub-classes from one to five respectively.} \label{tab:test2}
           \begin{subtable}{0.5\linewidth}
               \centering
               \begin{tabular}{cccccc} %
            \toprule
            \multirow{2}{*}{number}&SSL\cite{self2019supervised} &  \multicolumn{2}{c}{RotNet\cite{rot2018unsupervised}}& \multicolumn{2}{c}{Proposed} \\\cline{2-6}
                      &KL  &  MAX &	ODIN    &   MAX &   ODIN     \\
            \hline
            1& 0.91  &	0.91 &	\textbf{0.92} & 	-  & -        \\
            2& 0.91  &	0.89 &	0.89 &\textbf{0.97} &	\textbf{0.97} \\
            3&0.86   &	0.87 &	0.89 & \textbf{0.95} &	0.94  \\
            4& 0.82  &	0.87 &	0.87 & 0.93 &	\textbf{0.94} \\
            5&0.79   &	0.84 &	0.84 & \textbf{0.94} &	0.93 \\
            \toprule
            \end{tabular}
               \caption{MNIST dataset}
           \end{subtable}%
           \begin{subtable}{0.5\linewidth}
               \centering
            \begin{tabular}{cccccc} %
            \toprule
            \multirow{2}{*}{number}   & SSL\cite{self2019supervised} & \multicolumn{2}{c}{RotNet\cite{rot2018unsupervised}}    & \multicolumn{2}{c}{Proposed} \\ \cline{2-6}
                &KL       &  MAX &	ODIN    &   MAX &   ODIN           \\ 
            \hline
            1   &  \textbf{0.73}    &	0.62 &	0.66 &   -       &   -        \\
            2   &   0.76   &	0.72 &	0.72 &   \textbf{0.82}    &	0.81     \\
            3   &   0.66   &	0.64 &	0.61 &   \textbf{0.75}    &	0.72      \\
            4   &   0.56   &	0.60 &	0.55 &  \textbf{0.83}    &	0.76     \\
            5   &   0.54   &	0.57 &	0.55 &   \textbf{0.83}    &	0.79     \\
            \toprule
            \end{tabular} 
                \caption{CIFAR-10 dataset}
           \end{subtable}
\end{table*}
 
\section{Experimental Evaluation}  
\label{expresults}
   In our experimental evaluations, we use MNIST, CIFAR-10, CIFAR-100, and Imagenet. We use network constructed by two convolutional layers and two fully connected layers for MNIST testing. 
    We use Wide Residual Network(WRN) model \cite{wrn} which shows the state-of-the-art performance in classification task in CIFAR-10 data set. The parameters for depth and width of the model were selected as 28. We train the network with 0.1 learning rate at start and scaled down with cosine learning rate schedule\cite{cosine} using stochastic gradient descent with momentum at 0.9 during 200 epochs. We train the network with batches of size 250 on each in parallel on two TITAN Xp. 
    In SSL\cite{self2019supervised}, they select 30 classes out of 1000 classes so that there is no obvious overlap each other, unlike classes such as 'bee' and 'honeycomb'. We use pick randomly 10 classes out of 30 classes which is made up of 'airliner', 'american alligator', 'dragonfly', 'grand piano', 'hotdog', 'nail', 'snowmobile', 'stingray', 'soccer ball', 'tank'. For ImageNet, network is chosen to be residual network\cite{resnet} with self-attention module\cite{woo2018cbam} which is based on residual learning for high-resolution images. The number of layers is selected as 18 because we train on small number of classes. Augmentation of random crops and resize to 224 by 224 and random horizontal flips are applied. We train the network with learning rate initialized 0.1 and scaled down with 0.8 factor every 10 epochs using stochastic gradient descent with momentum at 0.9 during 200 epochs. We train the network with batches of size 250 on each in parallel on two TITAN Xp. 
    
    Table. \ref{expsetting} summarizes our proposed anomaly detection evaluation setup using MNIST, CIFAR-10, and ImageNet datasets. Our evaluation consists of five tests varying structure of normal and anomal data. Test 1 is traditional evaluation scheme of previous anomaly detection methods. We add four more comprehensive tests. 
 We use detection scores for each test following two methods\cite{msp}, \cite{odin}. In Hendrycks et al. \cite{msp}, they propose baseline method for out-of-distribution(OOD) detection claiming that well-trained network tends to give higher softMAX value to in-distribution data than out-of-distribution data. Similarly, Liang et al. \cite{odin} noted as ODIN add small perturbation to input image and scale the temperature showing effectiveness in OOD detection. We select both  measurements (noted as MAX \cite{msp}, ODIN \cite{odin}).
    For assigning the detection scores to SSL\cite{self2019supervised}, we follow the score function used in their paper in all of results tables. They measure the difference between softmax prediction of test sample and uniform distribution by KL divergence.   
    Area Under the Receiver-Operator-Characteristics(ROC) which indicates the ability of binary classification has been used as useful metric for evaluation the performance on many related works of anomaly detection tasks\cite{geo2018transform}, \cite{perera2019ocgan}, \cite{latent2019autoregressive}. We also compare the performance of our method using under the area of ROC curve noted as AUROC. 


\subsection{Test 1: Normal of Single Class, Anomaly of Multiple Classes}
\label{test1}
In the most of previous work, they have usually chosen one class as normal and the rest of classes as anomaly. However, normal data composed of only single class may make the task too easy and this set up would not be enough to reflect real situation.     
Table \ref{prevmnist} and \ref{prevcifar} show evaluation comparison with prior methods.
    We add more comprehensive results with this experiments, an algorithm 'Jigsaw' that the network is trained to predict order of shuffled images. 
Most algorithms in the table perform well with MNIST and CIFAR-10 data set, because the distribution of normal is relatively simple and easy to describe. 
SSL\cite{self2019supervised} and OCGAN\cite{perera2019ocgan} performs best on CIFAR-10 and MNIST, respectively.
Detecting diverse anomaly samples from single normal class of MNIST is kind of easy task achieving maximum 97.5\% of detection accuracy. In order for meaningful evaluation, we need extended evaluation scheme that we propose in the following evaluations. 

\subsection{Test 2: Normal of Multiple Classes, Anomaly of Multiple Classes} 
\label{test2} 
\begin{figure}[!b]
    \centering 
        \includegraphics[width=1\linewidth]{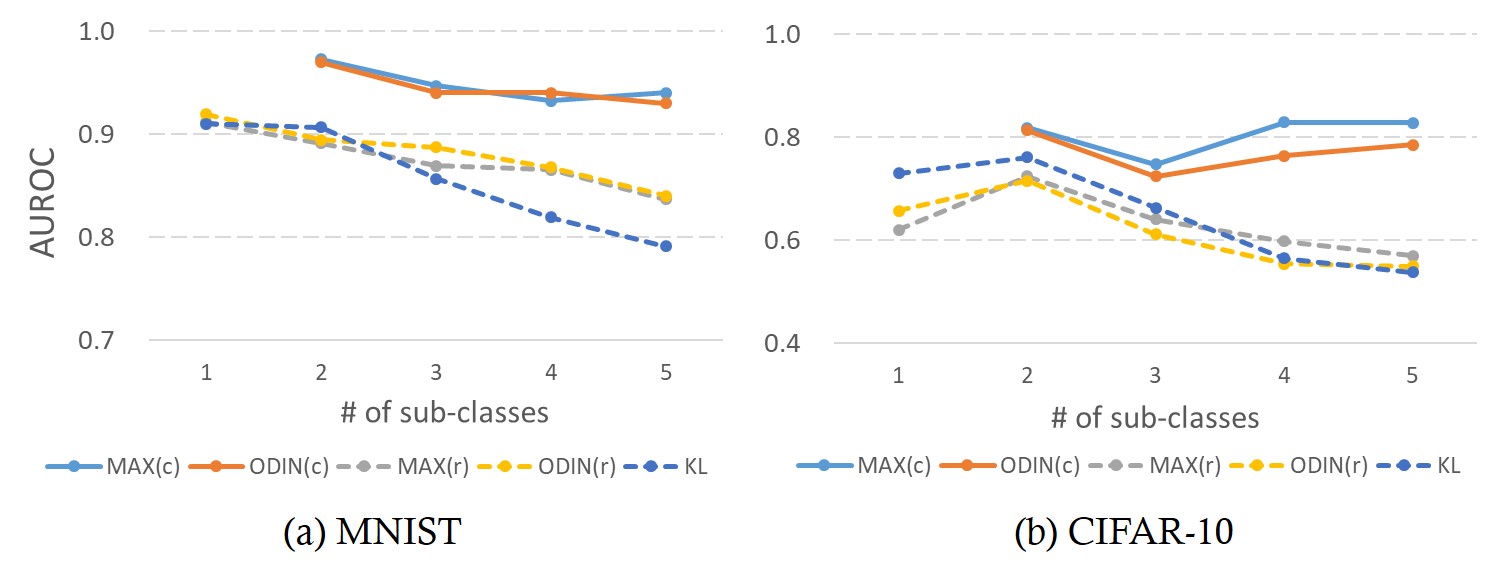} 
    \caption{AUROC scores as increasing the number of sub-classes in MNIST(left), CIFAR-10(right). Sky green and orange solid line show result for proposed method, grey and yellows dot line for rotation method and blue dot line show result for SSL\cite{self2019supervised}. }
    \label{inc}
\end{figure}

In our Test 2, extending traditional evaluation scheme, we test how the complexity of normal data affects anomaly detection performance by increasing the number of normal classes with MNIST and CIFAR-10 data set with fixed anomaly classes.
First five classes of both data sets are used as normal and remaining five classes are used as anomaly. Anomaly classes are fixed but the number of classes in normal data is increased from 1 to 5 in each experiment. For example, MNIST dataset, we use digit 1 as normal and digits 5,6,7,8,9 as anomaly in the first experiment. Next, we use digit 1,2 as normal and digit 5,6,7,8,9 as anomaly and so on. 
We compare our proposed scheme with state of the art method \cite{self2019supervised} and similar framework \cite{rot2018unsupervised} which is trained by predicting the angle of rotation.
Table \ref{tab:test2} summarize test results. 
In both results, proposed method shows almost similar AUROC as the number of normal classes increases.
However, the AUROC of SSL\cite{self2019supervised} and RotNet\cite{rot2018unsupervised} decreases losing its description ability on the normal data.
Note that SSL\cite{self2019supervised} is best performing method in Test 1 CIFAR-10 dataset.
We expected that better algorithm will maintain the anomaly detection performance regardless of the characteristic of normal dataset for same anomaly dataset. 
Figure \ref{inc} clearly shows that proposed method has such tendency.

\begin{table*}[!t]
           \centering
           \captionsetup[subtable]{position = below}
          \captionsetup[table]{position=top}
           \caption{ Test2: AUROC scores corresponding to the number of sub-classes included by normal set. (a), (b) For MNIST and CIFAR-10 dataset, AUROC score is calculated as increasing the number of sub-classes from one to five respectively.} \label{tab:test2}
           \begin{subtable}{0.5\linewidth}
               \centering
               \begin{tabular}{cccccc} %
            \toprule
            \multirow{2}{*}{number}&SSL\cite{self2019supervised} &  \multicolumn{2}{c}{RotNet\cite{rot2018unsupervised}}& \multicolumn{2}{c}{Proposed} \\\cline{2-6}
                      &KL  &  MAX &	ODIN    &   MAX &   ODIN     \\
            \hline
            1& 0.91  &	0.91 &	\textbf{0.92} & 	-  & -        \\
            2& 0.91  &	0.89 &	0.89 &\textbf{0.97} &	\textbf{0.97} \\
            3&0.86   &	0.87 &	0.89 & \textbf{0.95} &	0.94  \\
            4& 0.82  &	0.87 &	0.87 & 0.93 &	\textbf{0.94} \\
            5&0.79   &	0.84 &	0.84 & \textbf{0.94} &	0.93 \\
            \toprule
            \end{tabular}
               \caption{MNIST dataset}
           \end{subtable}%
           \begin{subtable}{0.5\linewidth}
               \centering
            \begin{tabular}{cccccc} %
            \toprule
            \multirow{2}{*}{number}   & SSL\cite{self2019supervised} & \multicolumn{2}{c}{RotNet\cite{rot2018unsupervised}}    & \multicolumn{2}{c}{Proposed} \\ \cline{2-6}
                &KL       &  MAX &	ODIN    &   MAX &   ODIN           \\ 
            \hline
            1   &  \textbf{0.73}    &	0.62 &	0.66 &   -       &   -        \\
            2   &   0.76   &	0.72 &	0.72 &   \textbf{0.82}    &	0.81     \\
            3   &   0.66   &	0.64 &	0.61 &   \textbf{0.75}    &	0.72      \\
            4   &   0.56   &	0.60 &	0.55 &  \textbf{0.83}    &	0.76     \\
            5   &   0.54   &	0.57 &	0.55 &   \textbf{0.83}    &	0.79     \\
            \toprule
            \end{tabular} 
                \caption{CIFAR-10 dataset}
           \end{subtable}
\end{table*}

\subsection{Test 3: Normal of Multiple Classes, Anomaly of Single Class} 
\label{test3}
    In this test, we select single class for anomaly and all remaining classes are used as normal maximizing the complexity of normal data set. This test is more similar to real world anomaly detection problems, where we have diverse types of normal data. 
In this experiment, we also consider each class of normal data as a semantic cluster that we expect. 
Table \ref{tab:test3} show test results on MNIST, CIFAR-10, and ImageNet dataset respectively. Our method outperforms SSL\cite{self2019supervised} in all three data sets. This results show how our approach is robust in diverse situations of normal and anomal data structure.
 
\begin{table*}[!t]
           \centering
           \captionsetup[subtable]{position = below}
          \captionsetup[table]{position=top}
           \caption{Test 3: Comparison of SSL\cite{self2019supervised} and proposed work in MNIST, CIFAR-10 and ImageNet dataset. AUROC from detection scores which calculated by by MAX\cite{msp} and ODIN\cite{odin}, KL divergence.}\label{tab:test3}
           \begin{subtable}{0.28\linewidth}
               \centering
               {\tiny
            \begin{tabular}{cccc} %
            \toprule
            \multirow{2}{*}{anomal} &SSL\cite{self2019supervised}   & \multicolumn{2}{c}{Proposed}               \\\cline{2-4}
                	                & KL   &  MAX                          &	ODIN     \\ 
            \hline
            0                       &\textbf{0.97}  &  \textbf{0.97}    &	\textbf{0.97} \\
            1                       &0.84  &  0.98              &\textbf{1.00}   \\
            2                       & 0.94 &  \textbf{0.97}      &	0.96     \\
            3                       & 0.78 &  \textbf{0.96}        &	0.95     \\
            4                       & 0.86 &  \textbf{0.94}     & \textbf{0.94}       \\
            5                       & 0.73 & \textbf{0.97}     &	0.93     \\ 
            6                       & 0.95 & \textbf{0.98}     &	0.97     \\
            7                       & 0.89 &  \textbf{0.96}    &	\textbf{0.96}    \\
            8                       & 0.81 &  \textbf{0.99}    &	\textbf{0.99}     \\
            9                       & 0.70 & \textbf{0.97}     &	0.96     \\
            \hline
            average&0.85	& \textbf{0.97} 	& 0.96 \\ 
            \toprule
            \end{tabular}}
               \caption{MNIST dataset }
           \end{subtable}%
          \hspace*{0.5em}
           \begin{subtable}{0.3\linewidth}
               \centering
               {\tiny
            \begin{tabular}{cccc} %
            \toprule
            \multirow{2}{*}{anomaly}   &  SSL\cite{self2019supervised}  & \multicolumn{2}{c}{Proposed} \\ \cline{2-4}
                    	               & KL    &  MAX                          &	ODIN     \\ 
            \hline
            airplane                   &  0.55 & 0.83                        &   \textbf{0.91}    \\
            automobile                 &0.39   & 0.70                         &   \textbf{0.78}     \\
            bird                       &0.74   &  0.79                       &   \textbf{0.85}     \\
            cat                        &0.77   & 0.79                       &\textbf{0.80}      \\
            deer                       &0.57   & 0.80                        &	\textbf{0.82}       \\
            dog                        & 0.71  & 0.76                      &	\textbf{0.78}       \\ 
            frog                       & 0.71  & 0.80                          &	\textbf{0.82}    \\
            horse                      & 0.43  & 0.80                         &   \textbf{0.81}     \\
            ship                       & 0.47  & 0.75                          &  \textbf{0.83}   \\
            truck                      & 0.49  & 0.74                          &   \textbf{0.85}     \\
            \hline
            average& 0.58  & 0.77 & \textbf{0.83}  \\
            \toprule
            \end{tabular}}
                \caption{CIFAR-10 datset}
           \end{subtable}
  \hspace*{0.5em}
           \begin{subtable}{0.34\linewidth}
               \centering
               {\tiny
            \begin{tabular}{cccc} %
\toprule
\multirow{2}{*}{anomaly}& SSL\cite{self2019supervised}  & \multicolumn{2}{c}{Proposed}      \\ \cline{2-4}
                        & KL   &  MAX                  &	ODIN     \\ 
\hline
centering  airliner     & 0.34 & \textbf{0.82}  & 0.76                       \\
american alligator      & 0.47 & 0.79  & \textbf{0.81}                      \\
dragonfly               & 0.60 & 0.72  & \textbf{0.73}                  \\
grand piano             & 0.60 & 0.86  & \textbf{0.88}                       \\
hotdog                  & 0.43 & 0.80  & \textbf{0.84}                      \\
nail                    & 0.69 & 0.79  & \textbf{0.81}                       \\
snowmobile              & 0.49 & 0.80  & \textbf{0.84}                      \\
stingray                & 0.54 & \textbf{0.76}  & \textbf{0.76}     \\
soccer ball             & 0.59 & \textbf{0.85}  & 0.85                      \\ 
tank                    & 0.42 & \textbf{0.80}  & \textbf{0.80}    \\
\hline
average& 0.52 & 0.80& \textbf{0.81}\\
\toprule
\end{tabular}}
                \caption{ImageNet datset}
           \end{subtable}
\end{table*}


\subsection{Test 4: Semantic Clustering with Labeled Data}
\label{test4} 
\begin{figure}[!b]
    \centering  
        \includegraphics[width=0.24\linewidth]{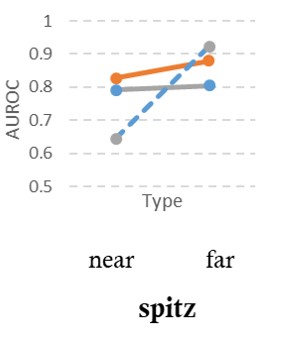} 
        \includegraphics[width=0.24\linewidth]{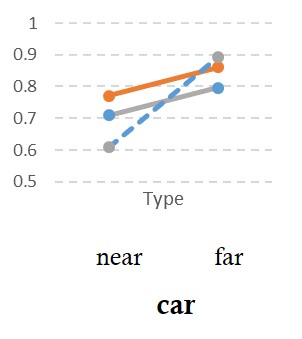}
        \includegraphics[width=0.24\linewidth]{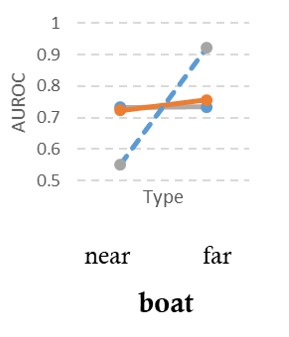} 
        \includegraphics[width=0.24\linewidth]{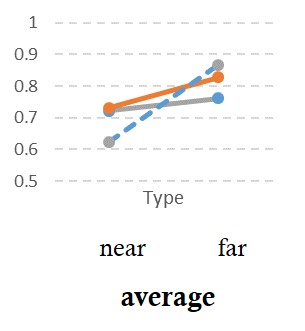}
    \caption{Test 4: Detection results with near and far anomaly using different types of normal sets. The results of our method are marked by orange(ODIN) and grey(MAX) solid line, SSL is marked by blue dot line. Proposed method is effective in detecting near (semantically closer) anomaly.}
    \label{comparetwoanomal}
\end{figure}

\begin{figure*}[ht]
\centering
    \includegraphics[width=\textwidth]{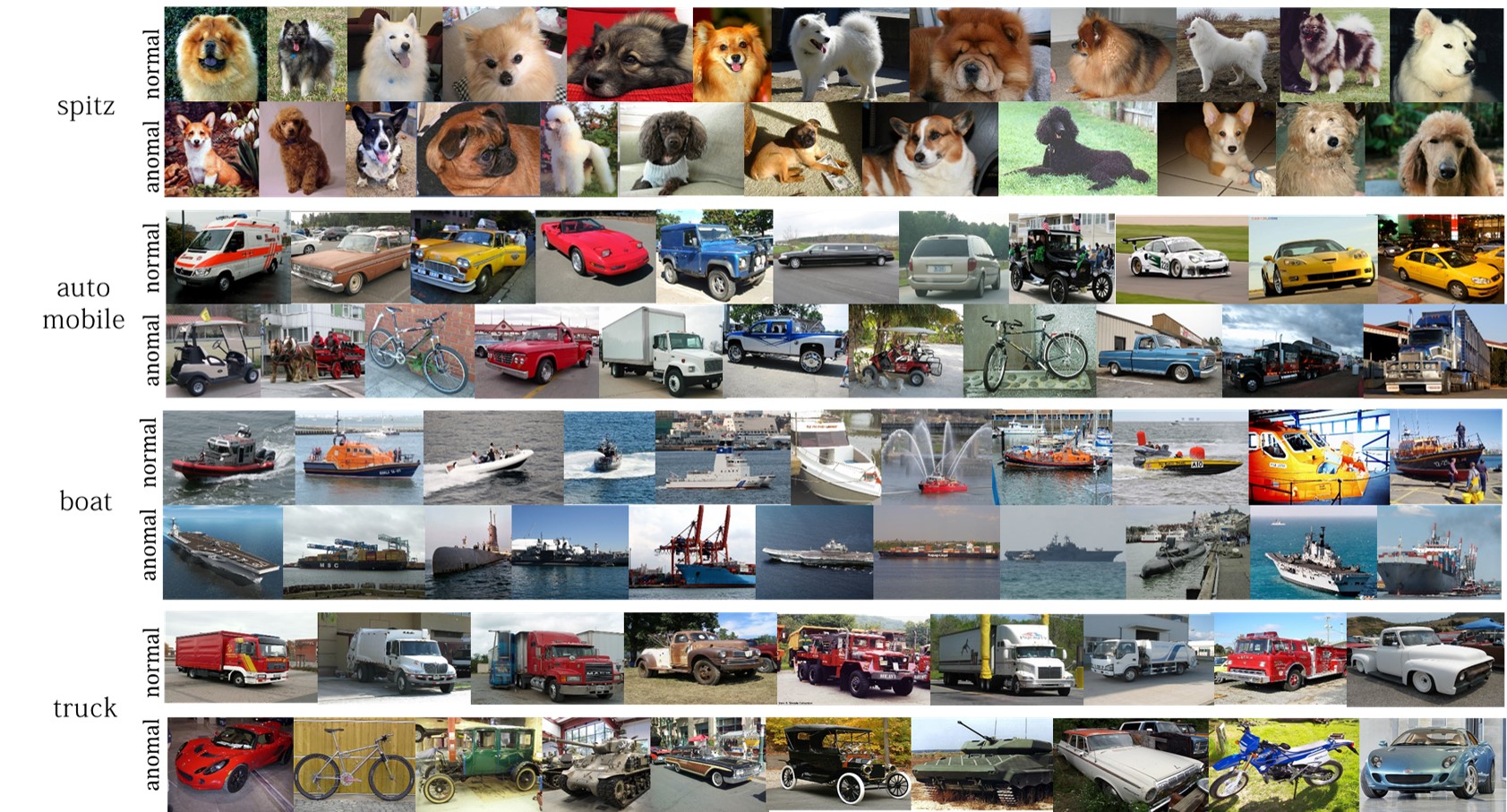}
    \caption{Examples of normal set and its correspondence anomalies(near) used in semantic clustering experiment.}
    \label{synset_example}
\end{figure*}

In test 4, we choose 'spitz', 'snake', 'car', 'boat', 'truck' and 'shop' synsets from ImageNet tree as respectively six normal sets. Sub-classes of each synset are considered as sub-clusters of normal data.  
For the rich simulation, we have conducted experiment under different semantic criteria that normal set has. 
For the anomaly, two different types (far, near) of anomal data are selected. Anomal(near) is visually very close to normal data and Anomal(far) is selected from less relevant category of normal data.
For example, for synset of auto mobile in figure \ref{synset_example}, all clusters in Anomal(near) of car types were not used in normal but belong to the transportation. 
'Spitz' set has 4 sub-clusters and different type of dogs consisted of 6 sub-classes are used as its Anomal(near). 'snake' set has 10 sub-classes and 5 other reptile sub-classes are used its Anomal(near). For 'car', 'boat' and 'truck' sets, 10, 3 and 5 sub-classes are used respectively and different types of transportation with its normal are used as its Anomal(near). 
Those five sets have shared Anomal(far) consisted of 4 sub-classes completely irrelevant to normal sets.
See first and third table \ref{synset}, first and second column depict the synset and sub-clusters respectively. In second table, we note the sub-clusters of Anomal(far) which consists of four classes from ImageNet.
For 'shop' set, 6 different classes of stores are used and 5 other types of place are used as its Anomal(near).
Our proposed method gives best average detection result for Anomal(near). SSL detects Anomal(far) relatively well because it has clear difference from normal data(see the table \ref{simanticclustering} and figure \ref{comparetwoanomal}).   
Similar to MNIST in \ref{test1}, sub-clusters in Anomal(far) is well separated from normal distribution. On the other hand, Anomal(near) has low separation from the normal distribution. 

\begin{table}[t]
\begin{center}
    \caption{ImageNet dataset for the evaluation of sub-clusters of normal data}
\label{synset} 
\begin{tabular}{cc} %
\toprule
   synset    &   sub-clusters\\
\hline
\multirow{1}{*}{\textbf{Normal(Spitz)}} &   Samoyed, Pomeranian, chow, keeshond \\ \cdashline{1-2}
\multirow{3}{*}{Anomal(near)} &   Brabancon griffon, Pembroke \\
                            & toy poodle, standardpoodle \\
                          &  Cardigan,  miniature poodle \\
\hline
\multirow{5}{*}{\textbf{Normal(snake)}}  & thunder snake, hognose snake\\ 
                            &king snake,night snake \\
                            &water snake, sea snake \\
                            &green snake, ringneck snake\\
                            &garter snake , vine snake \\ \cdashline{1-2} 
\multirow{3}{*}{Anomal(near)}  &green lizard,African crocodile\\
                            &boa constrictor,rock python\\
                            &Indian cobra\\
\midrule
\multirow{3}{*}{\textbf{Normal(car)}}  & ambulance, convertible, cab \\ 
                            &beach wagon ,Model T, jeep,minivan\\ 
                            &limousin, racer , sports car\\\cdashline{1-2} 
\multirow{2}{*}{Anomal(near)}   &golfcart,pickup,horse cart \\
                            &trailer truck,mountain bike \\ 
\hline
\multirow{1}{*}{\textbf{Normal(boat)}}  & fireboat, lifeboat,speed boat \\\cdashline{1-2} 
\multirow{1}{*}{Anomaly(near)}   &aircraft carrier, container ship,submarine\\  

\hline
\multirow{2}{*}{\textbf{Normal(truck)}}    & fire engine, garbage truck, pickup \\ 
                            &  tow truck, trailer truck\\\cdashline{1-2} 
\multirow{2}{*}{Anomal(near)} &beach wagon, Model T , moped\\
                            &mountain bike, sports car,tank\\ 
\toprule
\end{tabular} 
\\
\begin{tabular}{cc} %
\toprule
   synset    &   sub-clusters\\
\hline
\multirow{2}{*}{Anomal(far)} &  spider web, coral fungus\\  
                          &  cauliflower, custard apple\\ 
\toprule
\end{tabular} 
\end{center}
\end{table}

\begin{table}[t] 
\begin{center}
    \caption{Test 4: Results of semantic clustering experiment on six different normal data of ImageNet. We have explicitly decided the criteria for semantic clusters, first being an object and second being a place, as shown in the tables. In the left table, detection results are calculated about two different types of anomal set(near, far) and, in the right table, only near anomaly were used for detection.}
\label{simanticclustering} 
\begin{tabular}{ccccccc} %
\toprule
\multirow{3}{*}{normal}& \multicolumn{2}{c}{SSL\cite{self2019supervised}} &\multicolumn{4}{c}{Proposed}\\ \cline{2-7}
&\multicolumn{2}{c}{KL} & \multicolumn{2}{c}{MAX} &\multicolumn{2}{c}{ODIN}\\\cline{2-7} 
& near & far & near &far & near &  far\\
\hline
spitz       &  0.64     &\textbf{0.92}      &   0.79 &0.80 & \textbf{0.83}&0.88      \\
snake       &  0.59 &0.67       &  \textbf{0.71} &0.70  &  0.69 &\textbf{0.77 }  \\
car         &   0.61    & \textbf{0.89   }    & 0.71 &0.80& \textbf{ 0.77 }&0.86      \\
boat        &	0.55    &\textbf{0.92 }    &   \textbf{0.73}&0.73 &	0.72 &0.76    \\
truck       &    \textbf{0.73}&  \textbf{0.93  }      &0.66&0.75 &	0.70 &0.87          \\ 
\hline
average     & 0.62& \textbf{0.87} & 0.72 &0.76& \textbf{0.73}&0.83 \\
\toprule 
\end{tabular}
\quad
\begin{tabular}{cccc} %
\toprule
\multirow{2}{*}{normal} & {SSL\cite{self2019supervised}} &\multicolumn{2}{c}{Proposed}\\ \cline{2-4}
        &KL  & MSP & ODIN \\\cline{2-4}  
\hline
shop       &  0.64     &0.72      &  \textbf{0.80}   \\  
\toprule 
\end{tabular}
\end{center} 
\end{table}

\subsection{Test 5: Semantic Clustering with Unlabeled Data}
\label{test5} 
In previous four tests, we have evaluated anomaly detection performances based on explicitly labeled classes that are considered as good semantic clusters of normal data.   
However in real situation, we need to conduct appropriate semantic clustering of unlabeled normal data.
Semantic clustering is a challenging and critical task in unsupervised learning. Prior methods such as   
ISIF (Invariant and Spreading Instance Feature) \cite{embedding} show improved classification performance after the semantic clustering proving the ability of semantic separation of the method. 
ISIF propose unsupervised embedding learning which learns an embedded space preserving visual similarity or relation between categories of input images.
Because the embedded space reflects semantic information of given data, we employ the method proposed in ISIF for building semantic sub-clusters to conduct our test.
We again use ImageNet and CIFAR-100 which differ in complexity and resolution for making a normal set in different situation. 
For building the normal set, we make 'spitz', 'boat' and 'organism' sets by picking 4, 3 and 4 classes each from ImageNet. Their corresponding anomalies are 6 other types of dog, 3 other types of transport and 2 inanimate. In CIFAR-100 dataset, we make 'fish', 'large carnivores' and 'cat family' sets by picking 5, 5 and 3 classes respectively and their corresponding anomalies are 5 'aquatic mammals', 5 'omnivores and herbivores' and 2 'dog family'.

Note that we don't use any label information in this experiment. 
We train the embedded space following the training setting in ISIF. After training, we search K-nearest neighbor of a test input and measure the distance between the input and its neighbor in the feature space. We calculate cosine similarity and euclidean distance as the distance and these are used for a detection score to calculate AUROC score. For $K$ of 'K-nearest neighbor', we empirically choose 3 for 'spitz', 'fish', 'cat family' and 300 for 'organism', 'boat', 'carnivores'.
The AUROC scores of our proposed anomaly detection and SSL are listed in the table \ref{cluster_results}. Compared to SSL, our clustering based anomaly detection shows better performance on various types of unlabeled normal sets. KNN classification accuracy evaluates the semantic property used in each anomaly detection. Higher KNN accuracy means better semantic properties in the clustering step. Therefore, our proposed anomaly detection shows higher score when KNN accuracy is also higher. On the other hand, when KNN accuracy is lower, our clustering based anomaly detection performance is inevitably limited.

\begin{table}[!t]
\begin{center}
    \caption{Test 5: For the proposed anomaly detection using ISIF\cite{embedding}, AUROC scores is calculated over cosine similarity and euclidean distance and, for SSL\cite{self2019supervised}, KL divergence is used for calculating AUROC scores. The KNN accuracy evaluates the semantic ability of embedding model.}
\label{cluster_results} 
{\scriptsize
\begin{tabular}{cccccc} %
\toprule
\multirow{2}{*}{Dataset} & \multirow{2}{*}{Normal} & SSL\cite{self2019supervised}         &\multicolumn{2}{c}{Proposed}     & KNN   \\ \cline{3-5}
           &           & KL & Cosine &  Euclidean &  Accuracy\\ 
\hline
ImagetNet&spitz                  &\textbf{0.64}& 0.58 &0.59  & 0.57 \\
ImagetNet&boat                   &0.55 & 0.60  & \textbf{0.61} & 0.70 \\
ImagetNet&organism               &0.63 & \textbf{0.74} & \textbf{0.74} &0.81 \\ 
CIFAR-100&fish                   &0.60 & \textbf{0.68}  & \textbf{0.68}     &0.74 \\
CIFAR-100&carnivores             & 0.56 & \textbf{0.63}  & \textbf{0.63}   & 0.70\\ 
CIFAR-100&cat family             & \textbf{0.76} & 0.65  & 0.66 &0.72   \\  
\hline
&average & 0.62 & 0.65& \textbf{0.66} &0.70\\
\toprule
\end{tabular}
}
\end{center}
\end{table} 

\section{Conclusion and Future Works}
In this paper, we proposed a novel framework for anomaly detection which is learning the optimal feature space from semantic sub-clusters of normal data.  
We demonstrated the validity of our work on MNIST, CIFAR-10, CIFAR-100, ImagetNet, and unlabeled dataset by employing label classes or unsupervised embedding learning method which builds the feature space having semantic properties of data. 
We found that feature space learned through sub-clusters can dramatically improve the ability of anomaly detection. Our experimental results show outstanding performance even in high-dimensional real world data. 
   
{\small
\bibliographystyle{ieee_fullname}
\bibliography{egbib}
}

\end{document}